\documentclass{article} 
\usepackage[preprint]{style/colm2026_conference}
\usepackage[utf8]{inputenc}
\usepackage[T2A,T1]{fontenc}

\usepackage{microtype}
\usepackage{hyperref}
\usepackage{url}
\usepackage{booktabs}
\usepackage{multirow}
\usepackage{colortbl}
\usepackage{graphicx}
\usepackage{algorithm}
\usepackage{algpseudocodex}
\usepackage{calc}
\usepackage[dvipsnames]{xcolor}
\usepackage{menukeys} 
\usepackage{amsmath}
\usepackage{amssymb}
\usepackage[capitalize,noabbrev]{cleveref}

\usepackage{lineno}

\algnewcommand{\algorithmicand}{\textbf{and}}
\algnewcommand{\algorithmicor}{\textbf{or}}
\algnewcommand{\algorithmicnot}{\textbf{not}}
\algnewcommand{\algorithmicbreak}{\textbf{break}}

\newcommand{\seq}{\mathbf{x}}
\newcommand{\vocab}{\mathcal{V}}

\newcommand{\segs}[1]{\mathcal{S}(#1)}

\newcommand\vocabsize{n}

\newcommand{\nem}{N_{\text{em}}}

\newcommand{\mingrammi}{\mbox{MinGram-PP}}

\colorlet{tokgoodbg}{ForestGreen!22}
\colorlet{tokgoodbr}{ForestGreen!70!black}
\colorlet{tokmidbg}{Goldenrod!22}
\colorlet{tokmidbr}{Goldenrod!70!black}
\colorlet{tokbadbg}{BrickRed!22}
\colorlet{tokbadbr}{BrickRed!70!black}

\newmenucolortheme{tokgood}{named}{tokgoodbg}{tokgoodbr}{black}
\newmenucolortheme{tokmid} {named}{tokmidbg} {tokmidbr} {black}
\newmenucolortheme{tokbad} {named}{tokbadbg} {tokbadbr} {black}

\newcommand{\tokmethodbar}[3]{%
  \begin{tikzpicture}[baseline=-0.55ex,x=1cm,y=1cm]
    \pgfmathsetmacro{\tokgoodw}{1.6*(#1)}
    \pgfmathsetmacro{\tokmidw}{1.6*(#2)}
    \fill[tokgoodbg] (0,0) rectangle (\tokgoodw,0.14);
    \fill[tokmidbg] (\tokgoodw,0) rectangle ({\tokgoodw+\tokmidw},0.14);
    \fill[tokbadbg] ({\tokgoodw+\tokmidw},0) rectangle (1.6,0.14);
    \draw[black!20,line width=0.2pt] (0,0) rectangle (1.6,0.14);
  \end{tikzpicture}%
}

\ifdefined\tokenmacrosloaded \fi
\let\tokenmacrosloaded\relax

\IfFileExists{latexml.sty}{\usepackage{latexml}}{\newif\iflatexml\latexmlfalse}
\usepackage{xcolor}    

\colorlet{pretokbg}{white}
\colorlet{pretokbr}{black!45}
\colorlet{pretoktx}{black!75}

\iflatexml
\colorlet{tokbg}{black!12}
\colorlet{pretokbghtml}{black!4}   

\lxRequireResource[type=text/css,content={
.tok-key, .tok-key *, .pretok-key, .pretok-key * {
  font-family: ui-monospace, SFMono-Regular, Menlo, Consolas, monospace !important;
}
.tok-key, .pretok-key {
  font-size: 0.9em;
  border-radius: 4px;
  padding: 0.08em 0.32em;
  white-space: pre;
}
.tok-key {
  background-color: #e0e0e0;
  border: 1px solid #b5b5b5;
  border-bottom: 2px solid #9a9a9a;
  color: #111111;
}
.pretok-key {
  background-color: #f5f5f5;
  border: 1px dashed #9a9a9a;
  color: #444444;
}
}]{}

\makeatletter
\newcommand{\tk@box}[3]{
  \colorbox{#1}{\lxAddClass{#2}\ttfamily#3}%
}
\newcommand{\tk@list}[3]{
  \begingroup
  \def\tk@sep{}%
  \@for\tk@item:=#3\do{%
    \tk@sep
    \tk@box{#1}{#2}{\tk@item}%
    \def\tk@sep{\,}%
  }%
  \endgroup
}
\newcommand{\tokens}[2][]{\tk@list{tokbg}{tok-key}{#2}}
\newcommand{\pretokens}[1]{\tk@list{pretokbghtml}{pretok-key}{#1}}
\makeatother

\else
\usepackage{menukeys}
\renewmenumacro{\keys}[,]{roundedkeys}
\changemenuelement{roundedkeys}{sep}{\hspace{0.1em}}
\tikzset{tw@roundedkeys@base/.append style={font=\ttfamily}}
\newcommand{\tokens}[2][]{%
  \begingroup
  \if\relax\detokenize{#1}\relax\else
    \changemenucolortheme{roundedkeys}{#1}%
  \fi
  \keys{#2}%
  \endgroup
}

\copymenustyle{pretokkeys}{roundedkeys}
\changemenuelement{pretokkeys}{sep}{\hspace{0.3em}}
\tikzset{tw@pretokkeys@base/.append style={font=\ttfamily,dashed}}
\newmenucolortheme{pretok}{named}{pretokbg}{pretokbr}{pretoktx}
\changemenucolortheme{pretokkeys}{pretok}
\newmenumacro{\pretokens}[,]{pretokkeys}
\fi

\newcommand{\tsp}{\textvisiblespace}

\definecolor{darkblue}{rgb}{0, 0, 0.5}
\hypersetup{colorlinks=true, citecolor=darkblue, linkcolor=darkblue, urlcolor=darkblue}

\title{MinGram: A Minimalist Unigram Tokenizer with High Compression and Competitive Morphological Alignment}

\author{Sander Land \\
Writer, Inc.
\texttt{sander@writer.com} \\
}

\begin{document}

\ifcolmsubmission
\linenumbers
\fi

\maketitle

\begin{abstract}

The Unigram tokenizer uses an elegant representation which makes it straightforward to edit vocabularies, but its training is comparatively heavy and complex.
We introduce MinGram (Minimalist Unigram), which keeps the token-list representation but simplifies training using a BPE-derived seed vocabulary, Hard EM on a minimum-token path, and a single flat score-pruning step. This removes the suffix array, the forward-backward pass, and the iterative prune loop, leaving a procedure that requires little beyond tokenizer inference itself.
By making token count the primary objective and using a Unigram score only as a tiebreak, MinGram keeps the compression of pure token-count methods while retaining much of the morphological alignment and downstream quality of probabilistic ones.
Across six languages, MinGram compresses better than both BPE and standard Unigram, and a compression-oriented variant matches the strongest token-count compressors while retaining substantially higher morphological alignment. In controlled downstream language-model training, Unigram-family tokenizers, with MinGram among the best, consistently beat BPE in bits-per-byte.
\\
\href{https://github.com/sanderland/script_tok/}{
\raisebox{-0.15\height}{\includegraphics[width=0.3cm]{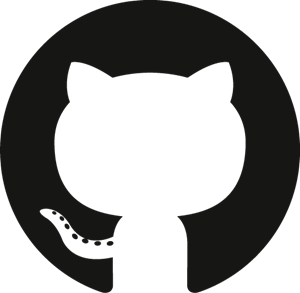}}
\texttt{\small\,github.com/sanderland/script\_tok}
}
\end{abstract}

\section{Introduction}

Tokenization sits at the entry point of almost every language model.
Nearly all modern language models use subword tokenization based on BPE~\citep{sennrich-etal-2016-neural},
 which greedily merges frequent adjacent pairs with a simple and well-understood algorithm.
It follows surface frequency without an explicit linguistic model, and recent work documents failure modes including undertrained tokens, poor morphological alignment, and vocabulary inefficiency~\citep{land-bartolo-2024-fishing, chizhov2024bpegetspickyefficient,asgari2025morphbpemorphoawaretokenizerbridging}.
The UnigramLM tokenizer~\citep[][hereafter `Unigram']{kudo-2018-subword} instead assigns probabilities to a vocabulary of pieces and defines a language model over segmentations. It is often reported to produce more morphologically plausible tokens~\citep{bostrom-durrett-2020-byte,vemula-etal-2025-rethinking}.  Compression, an important factor in the cost of training and using large language models, shows mixed results in comparisons with BPE~\citep{bostrom-durrett-2020-byte,land2025piecesdoesunigramtokenization}.

Unigram tokenization is attractive in part because of its representation as a simple list of tokens with scores: dropping tokens, pruning low-value entries, or combining vocabularies amounts to editing that list and re-normalizing scores. In BPE, by contrast, the vocabulary is tied to an ordered merge list, so removing or repairing tokens can require reasoning about downstream merge dependencies, as in recent work on BPE vocabulary cleanup~\citep{chizhov2024bpegetspickyefficient}.
The difficulty lies instead in Unigram training. Standard Unigram starts from suffix-array substring mining, estimates probabilities with forward-backward marginal EM over the segmentation lattice, and reaches the target vocabulary size through loss estimates in iterative pruning~\citep{kudo-richardson-2018-sentencepiece}.
This paper continues a line of work asking how the different components of Unigram training affect the properties of the resulting tokenizer. Prior work showed that the iterative pruning schedule and several SentencePiece defaults can be simplified with little loss, and sometimes with better compression~\citep{land2025piecesdoesunigramtokenization}. MinGram applies the same question to the remaining heavy components. Can we make the addition of a Unigram-style tokenizer to existing BPE-based implementations as simple as possible,
 and how can we optimize compression while maintaining some of the attractive morphological properties of Unigram?


We introduce \textbf{MinGram}, a minimalist Unigram tokenizer that keeps the token-list representation but
changes the focus towards compression and simplicity. It replaces the three main components of standard Unigram training:
\begin{enumerate}
    \item marginal EM is replaced by Hard EM on a minimum-token path
    \item iterative pruning is replaced by a single simple score-based pruning step
    \item suffix-array initialization is replaced by a BPE-derived seed vocabulary
\end{enumerate}

It keeps a learned Unigram score as a tiebreak among equally short segmentations, avoiding the pure least-token rule that can hurt morphology~\citep{schmidt-etal-2024-tokenization,uzan-etal-2024-greed}.
This places MinGram on a new point in the compression--morphology Pareto frontier, with strong compression combined with moderate morphological alignment.
The resulting tokenizer training requires little more than an implementation of its inference process,
 and in particular side-steps the need for lattices, forward-backward passes, loss-estimators in pruning, or substring mining.

\section{Background and Related Work}

\paragraph{BPE and UnigramLM.}
Byte-Pair Encoding \citep[BPE;][]{sennrich-etal-2016-neural} iteratively merges the most frequent adjacent symbol pair until the vocabulary reaches a target size.
The Unigram tokenizer~\citep{kudo-2018-subword,kudo-richardson-2018-sentencepiece} treats tokenization as probabilistic inference. A vocabulary $\vocab$ with token probabilities $p(\cdot)$ defines a distribution over segmentations $s = (s_1, \ldots, s_k)$ of a sequence $\seq$ via $p(s \mid \seq) \propto \prod_i p(s_i)$. Training starts from a large candidate vocabulary and alternates EM updates of $p$ with pruning steps that remove low-utility tokens, until $|\vocab| = \vocabsize$.
For a more detailed treatment of the implementation and mathematical background, see~\citet{land2025piecesdoesunigramtokenization} and~\citet{meister2026unigramlm}, respectively.

\paragraph{Adapting BPE.}
A line of recent work modifies BPE to optimize various aspects.
\citet{sennrich-etal-2017-university} recommend dropping rare subword units using training and test-time filtering.
ScaffoldBPE~\citep{scaffold_bpe} and PickyBPE~\citep{chizhov2024bpegetspickyefficient} implement a removal rule based on relative rather than absolute counts.
Several recent methods relax pretokenization to allow multi-word tokens and improve compression,
including SuperBPE~\citep{liu2025superbpe} and BoundlessBPE~\citep{schmidt2025boundless}. 
Finally, recent work by~\citet{foroutan2025parityaware} introduces a parity-aware variant that changes the merge objective to improve cross-lingual compression fairness.
These interventions are effective but often introduce complex workarounds due to the merge-list-based representation and application in BPE.
MinGram is in part motivated by these recent advances, as many such modifications are easier in the Unigram representation.

\paragraph{Compression and morphology.} Two common intrinsic measures of tokenizers are compression (as measured by e.g. tokens per character on held-out text) and morphological alignment, the extent to which a tokenizer's segmentations reflect a language's morphological structure. Tokenizer choice can affect downstream language-model performance,
and compression has been linked to translation performance~\citep{galle-2019-investigating}, but
such effects have not been consistent in more recent studies~\citep{schmidt-etal-2024-tokenization,lotz-etal-2025-beyond}.
The downstream evidence for morphological alignment specifically is also mixed~\citep{arnett2025evaluatingmorphologicalalignmenttokenizers,arnett-bergen-2025-language}.

\paragraph{Minimum-token inference.}
A separate line of work modifies tokenizer inference independently of training, motivated by tokenizer inference speed and model inference costs.
\citet{uzan-etal-2024-greed} evaluate different inference algorithms, including greedy variants and least tokens,
 and find that a pure least-token rule \emph{lowers} morphological alignment relative to greedy decoding. 
\citet{schmidt-etal-2024-tokenization} introduce PathPiece, a tokenizer which performs exact minimum-token inference and also trains vocabularies for this objective, but find that this objective does not consistently improve downstream performance.
Optimizing compression during training is a difficult problem, but ConvexTok~\citep{tempus2026tokenisationconvexrelaxations} recently formulated vocabulary construction as a linear program, yielding tokenizers with near-optimal compression.

\begin{algorithm*}[t]
\caption{MinGram training}
\label{alg:mingram}
\begin{algorithmic}[1]
\Require corpus $C$, target vocabulary size $\vocabsize$, overshoot factor $f$, EM iterations $\nem$
\State Train BPE on $C$ with vocabulary size $\lceil f \cdot \vocabsize \rceil$
\State Initialize $\vocab$ and $\log p$ from the BPE vocabulary and token frequencies
\For{$i = 1, \ldots, \nem$}
    \State Set token counts to zero
    \For{each sequence $\seq \in C$}
        \State Encode $\seq$ as $s^* = \arg\min_{s \in \segs{\seq}} \left( |s| - \delta \sum_j \log p(s_j) \right)$
        \State Add the tokens in $s^*$ to the corpus counts
    \EndFor
    \State Re-estimate $\log p(t)$ using normalized token frequencies
\EndFor
\State Prune to $|\vocab| = \vocabsize$ by lowest $\log p(t)$, preserving atomic tokens
\end{algorithmic}
\end{algorithm*}

\section{MinGram}

MinGram keeps the Unigram representation: a vocabulary with one score per token. It changes the inference and training procedure around that representation. Standard Unigram starts from a suffix-array candidate set, estimates token probabilities with marginal EM over the segmentation lattice, and reaches the target vocabulary size through iterative pruning. MinGram replaces these steps, aiming for simplicity and high compression: Hard EM on a minimum-token path, a BPE-derived seed vocabulary, and one flat score-pruning step.

\Cref{alg:mingram} summarizes the full training procedure. The implementation follows this procedure directly, with deterministic tie-breaking and atomic-token preservation handled in the encoder and pruning routines.

\subsection{Minimum-path inference}

Standard Unigram tokenization scores a segmentation using a probabilistic model over segmentations.
A common word can be split as \emph{root}+\emph{s} if the split has higher total probability than the whole-word token, even though it uses one more token. The same issue appears in marginal EM: probability mass is assigned across the full segmentation lattice, while compression depends on the paths that determine token count.

MinGram makes token count the primary decoding objective:
\begin{equation}\label{eq:mingram-objective}
    s^*(\seq) = \arg\min_{s \in \segs{\seq}}
    \left( |s| - \delta \sum_i \log p(s_i) \right),
\end{equation}
where $\delta$ is chosen small enough that one additional token always costs more than any possible log-probability gain. Thus the decoder first minimizes token count, then uses the Unigram score to choose among equally short segmentations. For a fixed vocabulary, the primary objective gives the best compression available to that vocabulary. The secondary score preserves useful frequency information without allowing it to choose a longer segmentation.
In case of an exact tie (which is common with long spans of identical characters), our implementation prefers the segmentation with longest leading tokens.

\subsection{BPE-seeded initialization}

Standard Unigram begins from a large suffix-array candidate set, typically 1M substrings, then prunes it down to the target vocabulary size. This seed is broad but wasteful: the process is complex, and efficient suffix-array construction with tokenizer-specific filtering is not widely implemented, leading to sub-optimal alternatives such as common substrings.

MinGram instead initializes from a BPE vocabulary trained on the same corpus, similar to PathPiece and SaGe~\citep{schmidt-etal-2024-tokenization,yehezkel-pinter-2023-incorporating}.
We train BPE to size $\lceil f \cdot \vocabsize \rceil$, where $f$ is a small overshoot factor, and use its tokens as the initial candidate vocabulary. This gives a compact seed whose tokens have already survived a compression-oriented construction process. We initialize $\log p(t)$ from each BPE token's corpus frequency.

\subsection{Minimum-path training}

Given this seed vocabulary, MinGram estimates token scores with Hard EM under the same minimum-path objective used for decoding. In the E-step, each training sequence is segmented as $s^*$ from \Cref{eq:mingram-objective}, and counts are accumulated only along that path. In the M-step, token scores are updated by normalized log-frequency:
\[
    \log p(t) \gets \log\left(\frac{\mathrm{count}(t)}{\sum_{t'} \mathrm{count}(t')} \right)
\]
This removes the forward-backward pass used by marginal Unigram EM.
After the EM iterations, MinGram prunes once to the target vocabulary size by flat score pruning, following~\citet{land2025piecesdoesunigramtokenization}.
We find that $\nem=2$ suffices, and an outer loop to slowly prune is not required in the regime of relatively small $f$, as shown in \Cref{app:em-ablation}.

The default MinGram method keeps this pruning step deliberately simple: after the final Hard-EM pass,
it removes the lowest-score non-atomic tokens in one step.

We also evaluate a compression-oriented variant, \mingrammi{} (MinGram with \textbf{P}athPiece-style \textbf{P}runing), whose final pruning rule instead iteratively removes the 10\% of tokens whose deletion least increases the corpus token count \citep{schmidt-etal-2024-tokenization}.
This method is notably slower and more complex to train, but retains the simple inference algorithm.


\section{Experimental Setup}\label{sec:setup}

We compare both MinGram variants against a range of other tokenizers: BPE, standard Unigram, Unigram with Flat Score Pruning \citep[FSP,][]{land2025piecesdoesunigramtokenization}, PathPiece-BPE~\citep{schmidt-etal-2024-tokenization}, and ConvexTok~\citep{tempus2026tokenisationconvexrelaxations}. 
We also test variants of standard Unigram and FSP with BPE-based initialization.
\Cref{tab:methods} summarizes how these methods differ in seed vocabulary, decoding objective, and pruning rule.
All experiments use SCRIPT encoding with character-boundary constraints as the atomic alphabet~\citep{scriptbpe},
 and train tokenizers with 32{,}768 learned tokens in addition to the required atomic tokens.
 The main setting trains one tokenizer per language on a 5\,GB FineWeb/FineWeb-2 sample
 for English, German, Finnish, Russian, Arabic, and Korean, 
 and evaluates compression on held-out Goldfish corpora for the same languages~\citep{chang2026goldfishmonolinguallanguagemodels}.
All BPE-initialized methods use an overshoot factor close to optimal for compression,
 using a large initial vocabulary ($f=8$) for \mingrammi{} and PathPiece and a mostly BPE-derived vocabulary for other BPE-initialized methods ($f = 1.15$). 

For measuring morphological alignment, we follow MorphAlign \citep{stephen2025morphtokeval},
 measuring the IBM1 alignment score at threshold $0.01$, multiplied by $100$,
 which we refer to as the `MorphAlign Score', on English, German, and Finnish.

\begin{table}[tbh]
\centering
\small
\setlength{\tabcolsep}{6pt}
\begin{tabular}{llll}
\toprule
Method & Seed & Objective & Token selection \\
\midrule
BPE              & bottom-up      & n/a                  & greedy, frequency-based          \\
Unigram          & suffix array   & max-prob             & loss estimate, iterative         \\
Unigram-BPE-Init & BPE            & max-prob             & loss estimate, single-step$^{*}$ \\
FSP              & suffix array   & max-prob             & flat score, iterative            \\
FSP-BPE-Init     & BPE            & max-prob             & flat score, single-step$^{*}$    \\
PathPiece-BPE    & BPE            & min-token            & token count loss, iterative      \\
ConvexTok        & all substrings & min-token            & token count, global (LP)         \\
\textbf{MinGram}      & BPE       & min-token + tiebreak & flat score, single-step$^{*}$    \\
\textbf{\mingrammi{}} & BPE       & min-token + tiebreak & token count loss, iterative      \\
\bottomrule
\end{tabular}
\caption{Tokenizers compared in this paper. \emph{Seed} is the candidate vocabulary, \emph{Objective} what inference optimizes (\emph{min-token + tiebreak} breaks ties by Unigram score), and \emph{Token selection} how the vocabulary is chosen. \emph{token count loss} greedily drops the token that least increases token count; ConvexTok minimizes the same objective globally as a linear program (LP). Single-step$^{*}$ rules collapse to one pass at our small seed sizes but support iterative pruning (cf. \Cref{app:em-ablation}).}
\label{tab:methods}
\end{table}

\section{Intrinsic performance}
\label{sec:results}

\begin{figure*}[p]
    \centering
    \includegraphics[width=0.85\linewidth]{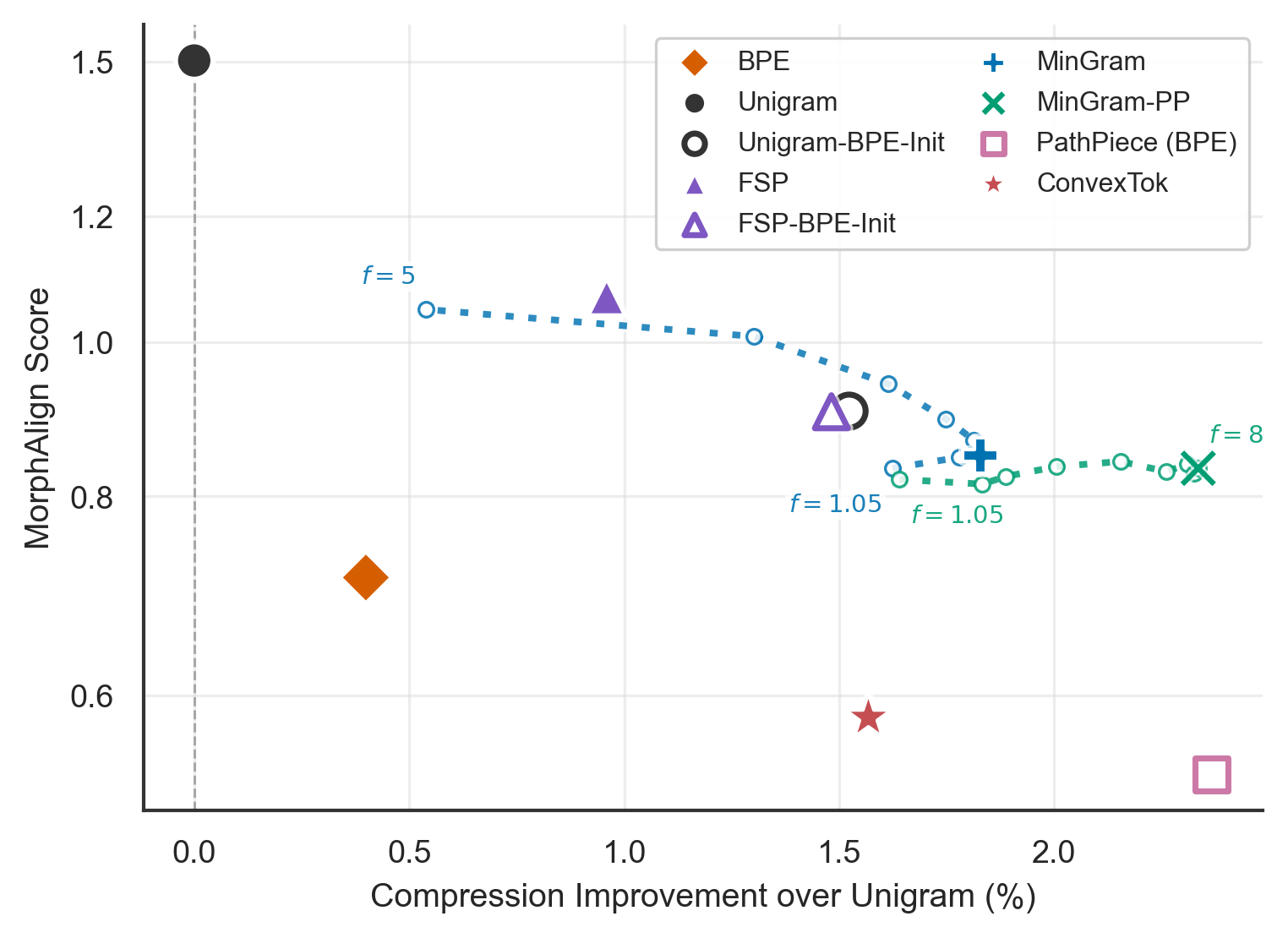}
    \caption{Compression improvement over Unigram versus MorphAlign Score. Compression is the arithmetic mean across the six compression languages. MorphAlign Score is the IBM1 alignment at threshold $0.01$ multiplied by $100$, with geometric mean across English, German, and Finnish. Higher is better on both axes. Single points show reference methods. The dotted curve traces MinGram variants as the overshoot factor $f$ varies. The highlighted default uses $f=1.15$ for MinGram and $f=8$ for \mingrammi{}.}
    \label{fig:lead}
\end{figure*}

\begin{table*}[p]
\centering
\small
\setlength{\tabcolsep}{4pt}
\begin{tabular}{lrrrrrrr}
\toprule
Method & English & German & Finnish & Russian & Arabic & Korean & Mean \\
\midrule
PathPiece\hspace{0pt}-BPE & \textbf{-1.18\%} & \textbf{-2.14\%} & \textbf{-2.87\%} & \textbf{-2.71\%} & \textbf{-2.59\%} & \textbf{-2.69\%} & \textbf{-2.37\%} \\
\mingrammi{} & -1.16\% & \underline{-2.10\%} & \underline{-2.85\%} & \underline{-2.66\%} & \underline{-2.56\%} & \underline{-2.67\%} & \underline{-2.33\%} \\
MinGram & -0.86\% & -1.49\% & -1.68\% & -2.37\% & -2.18\% & -2.40\% & -1.83\% \\
ConvexTok & \underline{-1.18\%} & -1.57\% & -1.76\% & -2.23\% & -1.89\% & -0.77\% & -1.57\% \\
Unigram\hspace{0pt}-BPE\hspace{0pt}-Init & -0.52\% & -1.22\% & -1.45\% & -2.11\% & -1.90\% & -1.95\% & -1.52\% \\
FSP\hspace{0pt}-BPE\hspace{0pt}-Init & -0.49\% & -1.17\% & -1.38\% & -2.07\% & -1.87\% & -1.91\% & -1.48\% \\
FSP & -0.33\% & -0.72\% & -1.16\% & -1.05\% & -1.10\% & -1.40\% & -0.96\% \\
BPE & +0.03\% & +0.46\% & +1.16\% & -1.25\% & -0.92\% & -1.88\% & -0.40\% \\
\bottomrule
\end{tabular}
\caption{Compression performance under monolingual FineWeb training: token count change (\%) relative to standard Unigram on Goldfish data; lower is better. Methods are sorted by mean compression. Bold marks best-in-column; underline marks second-best.}
\label{tab:main-compression}
\end{table*}

\begin{table}[p]
\centering
\small
\begin{tabular}{lrrrr}
\toprule
Method & English & German & Finnish & G. Mean \\
\midrule
Unigram & \textbf{0.98} & \textbf{1.72} & \textbf{2.02} & \textbf{1.50} \\
FSP & \underline{0.67} & \underline{1.14} & \underline{1.61} & \underline{1.07} \\
Unigram\hspace{0pt}-BPE\hspace{0pt}-Init & 0.66 & 0.89 & 1.27 & 0.91 \\
FSP\hspace{0pt}-BPE\hspace{0pt}-Init & 0.66 & 0.89 & 1.26 & 0.90 \\
MinGram & 0.57 & 0.87 & 1.23 & 0.85 \\
\mingrammi{} & 0.53 & 0.85 & 1.29 & 0.83 \\
BPE & 0.43 & 0.70 & 1.21 & 0.71 \\
ConvexTok & 0.36 & 0.59 & 0.94 & 0.58 \\
PathPiece\hspace{0pt}-BPE & 0.34 & 0.58 & 0.78 & 0.54 \\
\bottomrule
\end{tabular}
\caption{MorphAlign on the three UniMorph-evaluated languages, reported as IBM1 alignment at threshold $0.01$ multiplied by $100$; higher is better. Rows are sorted by G.\,Mean, the geometric mean across the three languages. Bold marks best-in-column; underline marks second-best.} 
\label{tab:main-morphalign}
\end{table}

\Cref{fig:lead} summarizes the main compression--morphology tradeoff.
Both MinGram variants show high compression, combined with moderate MorphAlign Score.
Notably, \mingrammi{} essentially ties with PathPiece for optimal compression,
while maintaining much higher MorphAlign Score compared to other compression-optimized methods such as ConvexTok and PathPiece.
Our proposed simpler MinGram algorithm is notably more sensitive to the choice of overshoot factor,
losing compression when using a larger initial vocabulary, where \mingrammi{} keeps gaining compression, 
saturating at $f=5\text{--}8$.
The MinGram $f$-trace shows that at higher overshoot factors, the compression gains reduce.
\Cref{app:fsweep,app:em-ablation} show this effect is in part due to the aggressive pruning,
which has detrimental effects at high $f$, but not at the default setting of $f=1.15$.

Unigram and FSP variants with BPE initialization show characteristics similar to MinGram,
losing a little compression performance due to the different objective, but clustering close together due to their similar vocabulary.
Overall the different variants show a subtle tradeoff, with a Pareto front mainly defined by default Unigram, FSP, and MinGram variants.
Additional results in \Cref{app:tiebreak} show the morphology gap between MinGram and PathPiece can be clearly attributed to the secondary term with a Unigram score objective.
In terms of these two metrics, BPE is Pareto-dominated by many Unigram-family methods in this intrinsic comparison, which all have both lower token count and higher MorphAlign Score.

\Cref{tab:main-compression} reports the per-language compression numbers, which are fairly consistent across languages.
\Cref{tab:main-morphalign} gives the MorphAlign breakdown. English and German show the clearest ordering: Unigram is highest, the compression-oriented reference methods and BPE are lowest, and MinGram lies between them, closer to the FSP and BPE-initialized variants than to either endpoint. 

\Cref{tab:tokenization-examples} illustrates the morphology side of the tradeoff with representative words, including the same compression-oriented reference methods. The examples match the aggregate pattern: MinGram often avoids BPE's most fragmented segmentations, but standard Unigram and FSP retain some cleaner morphology-oriented splits.

\section{Downstream language modeling}

\begin{table*}[t]
\centering
\small
\setlength{\tabcolsep}{6pt}
\begin{tabular}{lrlrlr}
\toprule
 & \multicolumn{2}{c}{bits/byte $\downarrow$} & \multicolumn{2}{c}{CORE $\uparrow$} & rare tokens $\downarrow$ \\
\cmidrule(lr){2-3}\cmidrule(lr){4-5}\cmidrule(l){6-6}
Method & bpb & $\Delta$ vs Best & CORE & $\Delta$ vs Best & count \\
\midrule
MinGram & \textbf{0.7123} & -- & 0.2658 & -0.82\% & \textbf{9} \\
Unigram\hspace{0pt}-BPE\hspace{0pt}-Init & \underline{0.7123} & -0.00\% & 0.2658 & -0.81\% & 26 \\
\mingrammi{} & 0.7125 & -0.03\%$^{*}$ & \textbf{0.2680} & -- & 57 \\
FSP\hspace{0pt}-BPE\hspace{0pt}-Init & 0.7126 & -0.04\%$^{**}$ & \underline{0.2664} & -0.61\% & 98 \\
Unigram & 0.7127 & -0.05\%$^{**}$ & 0.2625 & -2.05\%$^{*}$ & \underline{24} \\
FSP & 0.7127 & -0.07\%$^{**}$ & 0.2651 & -1.10\% & 49 \\
PathPiece\hspace{0pt}-BPE & 0.7131 & -0.12\%$^{**}$ & 0.2643 & -1.38\% & 59 \\
ConvexTok & 0.7132 & -0.13\%$^{**}$ & 0.2631 & -1.83\%$^{*}$ & 81 \\
BPE & 0.7138 & -0.22\%$^{**}$ & 0.2605 & -2.78\%$^{*}$ & 172 \\
\bottomrule
\end{tabular}

\caption{Downstream English language modeling. Rows are sorted by held-out bits-per-byte (bpb; lower is better). Each tokenizer was trained on FineWeb-English, and depth-24 nanochat models were trained on 5.86B tokens of ClimbMix with $n = 20$ seeds per method. We report bpb and DCLM CORE centered accuracy, with $\Delta$ relative to the best row for each metric; negative values are worse than best on both axes. Stars denote Welch's $t$-tests against that metric's best row ($^*p{<}.05$, $^{**}p{<}.001$). The rare tokens metric shows the number of tokens seen less than once every $10^{7}$ tokens.}
\label{tab:downstream}
\end{table*}

As a controlled downstream experiment, we train depth-24 nanochat \citep{nanochat} language models,
using the default setup of 5.86B tokens of the (primarily English) ClimbMix dataset \citep{diao2025nemotronclimbclusteringbasediterativedata} 
and measure performance in bits-per-byte (bpb) and the DCLM CORE score \citep{li2024datacomplm},
 following \citet{schmidt2026tokenizationsplittrees}.
We use 20 random seeds per method based on power calculations performed after 5 seeds per tokenizer.

Results in \Cref{tab:downstream} show MinGram variants among the best in bits-per-byte, with Unigram-family methods dominating the top of the ranking. ConvexTok and PathPiece-BPE fall clearly behind, and BPE is worst. This metric is also highly consistent across seeds, which makes the differences generally significant.

The CORE score shows high variability with low scores for many tasks, as models are undertrained for many of the benchmarks.
\mingrammi{} performs best, but only default Unigram, BPE and ConvexTok show significantly worse performance.
Larger models and more training data are required to better separate tokenizer performance.
Finally, we inspect the number of rare tokens in the models' training corpus, a proxy for under-trained vocabulary entries \citep{land-bartolo-2024-fishing}: MinGram has the fewest, suggesting the pruning strategy is less sensitive to outliers in training data.
We attribute the difference in the number of rare tokens between MinGram and PathPiece to PathPiece's objective (shared by \mingrammi{}), which keeps semi-rare tokens from being shattered into many fragments and retains them in the vocabulary.
For example, \mingrammi{} and PathPiece vocabularies include \tokens{ArticlePubMedGoogle},
which is represented as \tokens{Article,PubMed,Google} in default MinGram.
\Cref{app:raretokens} gives additional examples of rare tokens and their frequencies in the training corpus.

\begin{table*}[t]
\centering
\small
\caption{Qualitative tokenization examples for representative English, German, and Finnish words, including PathPiece-BPE and ConvexTok as reference methods. Vertical bars mark token boundaries. The color bar under each method header summarizes that method's boundary quality across all example words: green = no extra predicted boundaries, yellow = one extra, red = two or more.}
\label{tab:tokenization-examples}
\resizebox{\textwidth}{!}{
\begingroup
\setlength{\tabcolsep}{1.8pt}
\tiny
\begin{tabular}{@{}lllllll@{}}
\toprule
Gold & BPE & Unigram & FSP & MinGram & PathPiece-BPE & ConvexTok \\
 & \tokmethodbar{0.188}{0.562}{0.250} & \tokmethodbar{0.500}{0.438}{0.062} & \tokmethodbar{0.500}{0.500}{0.000} & \tokmethodbar{0.625}{0.375}{0.000} & \tokmethodbar{0.438}{0.500}{0.062} & \tokmethodbar{0.250}{0.625}{0.125} \\
\midrule
\multicolumn{7}{@{}l@{}}{\small\textbf{English}} \\
\tokens{out,spoken} & \tokens[tokgood]{out,spoken} & \tokens[tokbad]{outs,po,ken} & \tokens[tokmid]{out,spoke,n} & \tokens[tokgood]{out,spoken} & \tokens[tokbad]{outs,po,ken} & \tokens[tokgood]{out,spoken} \\
\addlinespace[0.2em]
\tokens{space,ship} & \tokens[tokbad]{sp,aces,hip} & \tokens[tokgood]{space,ship} & \tokens[tokgood]{space,ship} & \tokens[tokgood]{space,ship} & \tokens[tokgood]{space,ship} & \tokens[tokgood]{space,ship} \\
\addlinespace[0.2em]
\tokens{south,east} & \tokens[tokmid]{s,outheast} & \tokens[tokgood]{south,east} & \tokens[tokgood]{south,east} & \tokens[tokmid]{s,outh,east} & \tokens[tokgood]{south,east} & \tokens[tokmid]{so,uth,east} \\
\addlinespace[0.2em]
\tokens{colour,ed} & \tokens[tokmid]{col,oured} & \tokens[tokgood]{colour,ed} & \tokens[tokgood]{colour,ed} & \tokens[tokmid]{col,oured} & \tokens[tokgood]{colour,ed} & \tokens[tokgood]{colour,ed} \\
\addlinespace[0.2em]
\tokens{rough,neck} & \tokens[tokgood]{rough,neck} & \tokens[tokgood]{rough,neck} & \tokens[tokmid]{r,ough,neck} & \tokens[tokgood]{rough,neck} & \tokens[tokmid]{r,ough,neck} & \tokens[tokmid]{r,ough,neck} \\
\addlinespace[0.2em]
\tokens{shop,lift,er} & \tokens[tokmid]{shop,lif,ter} & \tokens[tokgood]{shop,lift,er} & \tokens[tokgood]{shop,lift,er} & \tokens[tokgood]{shop,lift,er} & \tokens[tokgood]{shop,lift,er} & \tokens[tokgood]{shop,lift,er} \\
\addlinespace[0.2em]
\tokens{ultra,modern} & \tokens[tokbad]{ult,ram,od,ern} & \tokens[tokmid]{ul,tra,modern} & \tokens[tokmid]{ul,tra,modern} & \tokens[tokmid]{ult,ra,modern} & \tokens[tokmid]{ul,tra,modern} & \tokens[tokmid]{ul,tra,modern} \\
\addlinespace[0.35em]
\multicolumn{7}{@{}l@{}}{\small\textbf{German}} \\
\tokens{Ach,tel,note} & \tokens[tokmid]{Ach,teln,ote} & \tokens[tokgood]{Ach,tel,note} & \tokens[tokmid]{Acht,el,note} & \tokens[tokgood]{Ach,tel,note} & \tokens[tokgood]{Ach,tel,note} & \tokens[tokmid]{A,chtel,note} \\
\addlinespace[0.2em]
\tokens{Nacht,falter} & \tokens[tokmid]{Nacht,f,alter} & \tokens[tokmid]{Nacht,falt,er} & \tokens[tokmid]{Nacht,falt,er} & \tokens[tokmid]{Nacht,falt,er} & \tokens[tokmid]{Nacht,f,alter} & \tokens[tokmid]{Nacht,f,alter} \\
\addlinespace[0.2em]
\tokens{be,dräng,en} & \tokens[tokbad]{bed,rän,gen} & \tokens[tokmid]{be,dränge,n} & \tokens[tokmid]{be,dr,ängen} & \tokens[tokgood]{be,drängen} & \tokens[tokmid]{be,dr,ängen} & \tokens[tokmid]{be,dr,ängen} \\
\addlinespace[0.2em]
\tokens{zer,kratz,en} & \tokens[tokbad]{z,erk,ratzen} & \tokens[tokmid]{zer,krat,zen} & \tokens[tokgood]{zer,kratz,en} & \tokens[tokgood]{zer,kratz,en} & \tokens[tokmid]{zer,kra,tzen} & \tokens[tokmid]{zer,kra,tzen} \\
\addlinespace[0.35em]
\multicolumn{7}{@{}l@{}}{\small\textbf{Finnish}} \\
\tokens{kestä,nyt} & \tokens[tokmid]{kest,änyt} & \tokens[tokmid]{ke,stä,nyt} & \tokens[tokmid]{kes,tänyt} & \tokens[tokgood]{kestä,nyt} & \tokens[tokmid]{kes,tänyt} & \tokens[tokmid]{kes,tänyt} \\
\addlinespace[0.2em]
\tokens{toista,isi,t} & \tokens[tokgood]{toista,isit} & \tokens[tokgood]{toista,isi,t} & \tokens[tokgood]{toista,isit} & \tokens[tokgood]{toista,isit} & \tokens[tokgood]{toista,isit} & \tokens[tokmid]{toista,i,sit} \\
\addlinespace[0.2em]
\tokens{virtaa,mme} & \tokens[tokmid]{vir,taa,mme} & \tokens[tokgood]{virtaa,mme} & \tokens[tokgood]{virtaa,mme} & \tokens[tokgood]{virtaa,mme} & \tokens[tokmid]{virta,amme} & \tokens[tokmid]{virta,amme} \\
\addlinespace[0.2em]
\tokens{matkusta,isi,mme} & \tokens[tokmid]{mat,kusta,isimme} & \tokens[tokmid]{matkust,a,isi,mme} & \tokens[tokgood]{matkusta,isimme} & \tokens[tokmid]{mat,kusta,isimme} & \tokens[tokgood]{matkusta,isimme} & \tokens[tokbad]{mat,kus,tai,simme} \\
\addlinespace[0.2em]
\tokens{muodosta,n} & \tokens[tokmid]{muod,osta,n} & \tokens[tokmid]{muodo,sta,n} & \tokens[tokmid]{muodo,sta,n} & \tokens[tokmid]{muod,osta,n} & \tokens[tokmid]{muodo,stan} & \tokens[tokbad]{mu,odo,stan} \\
\bottomrule
\end{tabular}
\endgroup}
\end{table*}

\section{Conclusion}

MinGram occupies a new point on the compression–morphology Pareto frontier. It compresses better than both BPE and standard Unigram while retaining moderate morphological alignment, and it does so with a procedure that needs little beyond tokenizer inference: no suffix array, forward-backward pass, or iterative pruning loop.
The default method trades a small amount of maximal compression for this simplicity, and is more sensitive than standard Unigram to the size of the initial vocabulary. Careful pruning reduces this sensitivity, but fully closing the compression gap requires the more complex iterative rule of \mingrammi{}.
That variant essentially matches PathPiece-BPE for the strongest compression in our comparison, yet retains more of the longer rare tokens, which suggests the pruning rule could be made more robust to outliers in the training data without sacrificing compression.
We recommend the default MinGram for most settings: it has the simplest training, is among the best on downstream bits-per-byte (better than \mingrammi{}), and produces the fewest rarely occurring tokens in a cross-corpus setting. \mingrammi{} is preferable when token-count compression is the dominant cost, for instance to shorten sequences and reduce serving cost at scale: its added complexity is confined to training, while inference is identical to MinGram and emits fewer tokens. It matches the strongest token-count compressors, trading a small bits-per-byte penalty and more rarely-used vocabulary entries for roughly 0.5 percentage points of additional compression.
Our downstream results also caution against optimizing token count alone: the methods that optimize token count without a probability score were among the worst in bits-per-byte, while tokenizers that retain a Unigram score, including the equally compression-strong \mingrammi{}, modeled text best. 
More broadly, this adds to the evidence that compression alone is the wrong objective for tokenizer design: a Unigram-style language-modeling score, even one used only to break ties in segmentation, measurably improves the resulting tokenizer and carries through to downstream modeling.

\section*{Limitations}

Our downstream evaluation is narrow. It uses a single small model (depth-24 nanochat), a single training corpus (ClimbMix), one vocabulary size, and a 5.86B-token budget, all in English. Compression differences were clearly resolved in bits-per-byte, but the DCLM CORE task scores were too noisy at this scale to separate most methods: only default Unigram, BPE, and ConvexTok were significantly worse than the best, while the remaining methods were statistically indistinguishable. The downstream ranking we report therefore rests on bits-per-byte rather than task performance, and both are likely to change at larger model and data scales.

Morphological alignment is evaluated only where UniMorph-based MorphAlign data are available: English, German, and Finnish. These are useful probes but do not cover the full range of scripts, segmentation conventions, or morphological systems in our six compression languages.


\bibliography{references}
\bibliographystyle{colm2026_conference}

\appendix
\crefalias{section}{appendix}

\onecolumn

\section{Effect of the BPE-derived initial vocabulary size}
\label{app:fsweep}

\Cref{fig:lead} shows MinGram's $f$-trace averaged across the MorphAlign-evaluated languages. This appendix examines the sensitivity to the overshoot factor $f$ defining the size of the BPE vocabulary in different tokenizers using a BPE-initialized vocabulary.

\begin{figure}[!ht]
    \centering
    \includegraphics[width=0.7\linewidth]{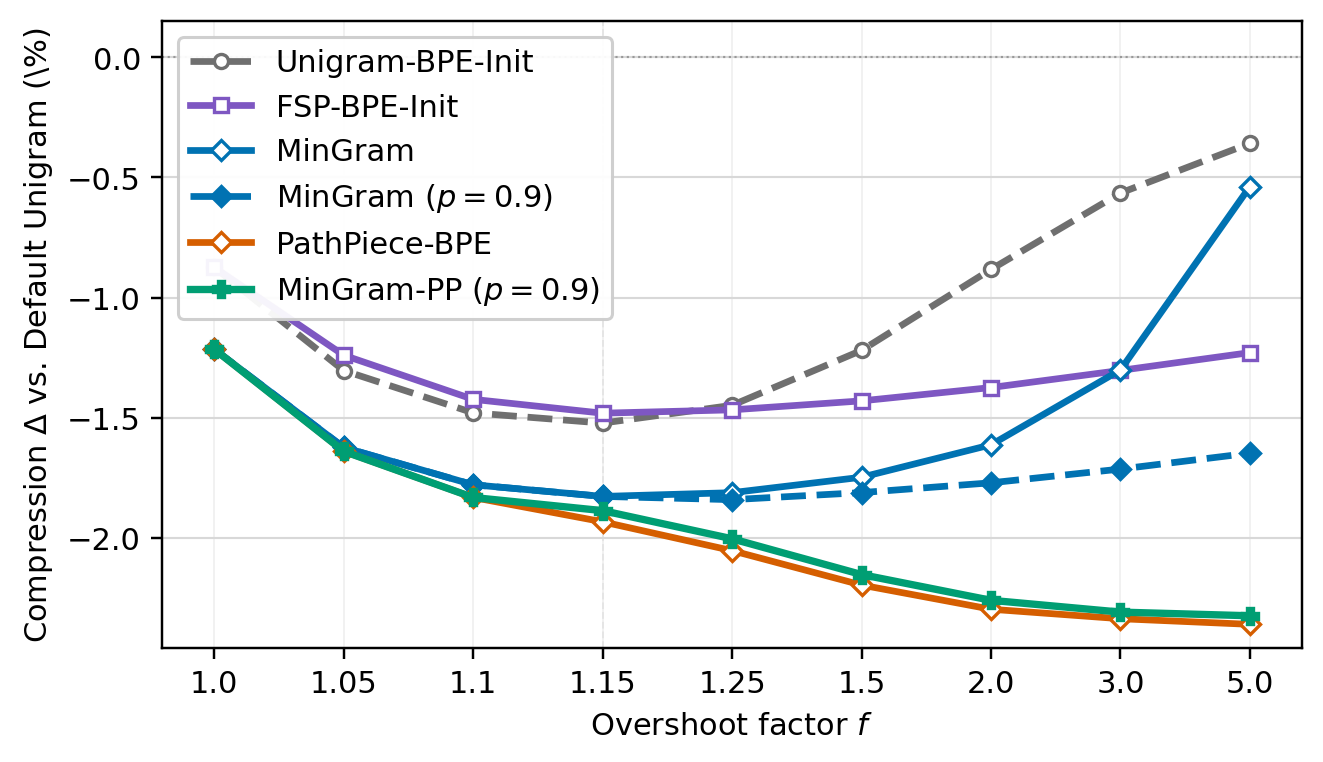}
    \caption{Compression sensitivity to overshoot factor $f$, measured as token-count change relative to Default Unigram (Goldfish data, mean across six languages). Lower is better. PathPiece-BPE is placed on the same BPE-seed vocabulary-size axis; dashed MinGram shows iterative pruning.}
    \label{fig:fsweep-compression}
\end{figure}

\Cref{fig:fsweep-compression} shows compression as a function of $f$ for the BPE-init methods, measured as percentage change relative to Default Unigram. Unigram-BPE-Init and FSP-BPE-Init coincide at $f = 1$ by construction; MinGram and PathPiece-BPE use the same BPE-seed size axis but differ in training and inference. Increasing $f$ first improves compression, with the six-language mean minimum at $f = 1.15$ for the main BPE-init Unigram-family methods. This point is the default used in the main intrinsic comparisons. At larger $f$, Unigram-BPE-Init moves back toward the Default Unigram baseline, while MinGram also weakens. Including a run with gradual pruning ($p=0.9$)
 shows that this high-$f$ loss is partly a one-step pruning effect.
 With a very large BPE seed, single-step score pruning must discard many competing long candidates at once, while iterative pruning recovers $>1$\,pp of compression at $f = 5$ by allowing the model to resegment between pruning steps, though its optimum remains at $f=1.15$.
The more careful compression-optimized methods, PathPiece and \mingrammi{}, benefit from larger initial vocabularies, saturating at around $f=5\text{--}8$.

\clearpage

\section{EM Iterations and Pruning Schedule}
\label{app:em-ablation}

This section motivates the MinGram defaults: $\nem = 2$ inner EM iterations and single-step pruning ($p = 0$).
It also investigates to what extent the compression difference between MinGram and \mingrammi{} can be closed by more careful pruning.

\Cref{tab:em-ablation} reports MorphAlign and compression as functions of $\nem \in \{0, 1, 2, 3, 4\}$ for MinGram at $f = 1.15$ on English, German, and Finnish, with PathPiece-BPE included as a minimum-token reference point.
$\nem = 0$ drops MorphAlign substantially because BPE-derived counts are a poor prior for score-based pruning. At least one EM pass is needed to re-estimate probabilities under the MinGram objective. One to three iterations are all reasonable choices; we use $\nem = 2$ as the default. PathPiece-BPE is more compression-oriented than these MinGram settings, but has lower MorphAlign even than the no-EM MinGram variant.

\Cref{tab:pruning-schedule} compares single-step pruning ($p = 0$) with iterative pruning ($p = 0.9$), both using MinGram's score-based criterion. This isolates the pruning schedule from the separate token-count rule of \mingrammi{}.  To make the comparison visible, the table shows the single-step mean and the iterative-minus-single-step delta over English, German, and Finnish. For $f \leq 1.15$, the candidate pool is small enough that both schedules effectively collapse to the same outcome, covering the default setting. Larger $f$ values recover some compression under iterative pruning, but require extra corpus tokenizations and do not improve MorphAlign. We therefore retain single-step pruning as the default.

\begin{table}[hb]
\centering
\small
\begin{tabular}{lrrrr}
\toprule
Setting & English & German & Finnish & Mean \\
\midrule
\multicolumn{5}{l}{\textit{MorphAlign Score}} \\
0 & 0.52 & 0.77 & 1.16 & 0.77 \\
1 & 0.56 & 0.87 & 1.21 & 0.84 \\
2 & 0.57 & 0.87 & 1.23 & 0.85 \\
3 & 0.57 & 0.88 & 1.23 & 0.85 \\
4 & 0.57 & 0.88 & 1.23 & 0.85 \\
PathPiece\hspace{0pt}-BPE & 0.34 & 0.58 & 0.78 & 0.54 \\
\midrule
\multicolumn{5}{l}{\textit{Compression $\Delta$ vs Unigram (\%, lower is better)}} \\
0 & -0.71\% & -1.03\% & -0.85\% & -0.86\% \\
1 & -0.84\% & -1.47\% & -1.55\% & -1.29\% \\
2 & -0.86\% & -1.49\% & -1.68\% & -1.34\% \\
3 & -0.86\% & -1.49\% & -1.71\% & -1.35\% \\
4 & -0.86\% & -1.49\% & -1.73\% & -1.36\% \\
PathPiece\hspace{0pt}-BPE & -1.18\% & -2.14\% & -2.87\% & -2.07\% \\
\bottomrule
\end{tabular}
\caption{MorphAlign and compression delta vs default Unigram as functions of EM iterations $\nem$ for MinGram at $f = 1.15$, with PathPiece-BPE as a reference. $\nem = 0$ uses BPE-derived counts directly with no EM.}
\label{tab:em-ablation}
\end{table}

\begin{table}[hb]
\centering
\small
\begin{tabular}{lrrrr}
\toprule
 & \multicolumn{2}{c}{MorphAlign Score $\times 100$} & \multicolumn{2}{c}{Compression $\Delta$ mean} \\
\cmidrule(lr){2-3}\cmidrule(lr){4-5}
$f$ & single-step & iter. $-$ single & single-step & iter. $-$ single \\
\midrule
1.1 & 0.85 & +0.00 & -1.23\% & +0.00 pp \\
1.15 & 0.85 & +0.00 & -1.34\% & +0.00 pp \\
1.25 & 0.87 & +0.00 & -1.34\% & -0.03 pp \\
1.5 & 0.89 & -0.01 & -1.32\% & -0.05 pp \\
2 & 0.94 & -0.03 & -1.25\% & -0.13 pp \\
3 & 1.01 & -0.07 & -1.10\% & -0.28 pp \\
5 & 1.05 & -0.09 & -0.61\% & -0.77 pp \\
\bottomrule
\end{tabular}
\caption{MorphAlign and compression delta vs default Unigram for single-step pruning at $\nem = 2$, with iterative pruning shown as a delta from single-step. Negative compression deltas mean iterative pruning compresses more.}
\label{tab:pruning-schedule}
\end{table}

\clearpage

\section{Effect of the Unigram Score Objective}
\label{app:tiebreak}

The comparison of \mingrammi{} with PathPiece suggests the effect of the small term with a Unigram score in the objective (\Cref{eq:mingram-objective}) is to increase morphological alignment without affecting compression.
To confirm and quantify this effect in the default MinGram algorithm, we compare the default Unigram log-probability score tiebreak against two controls while keeping both the primary minimum token count objective fixed.
\Cref{tab:tiebreak} shows that when replacing the token log-probability scores with random values, MorphAlign Score decreases to that of PathPiece. Furthermore, when we invert the term to prefer lower-probability paths, the MorphAlign Score decreases even more to the lowest among all methods.
Compression is identical across all three MinGram policies by construction because they all optimize the same minimum-token-count objective.
This result shows that the Unigram objective, despite being a secondary tie-breaker term, still has a large effect on the chosen segmentation.

\begin{table}[!ht]
\centering
\small
\begin{tabular}{lrrrr|rr}
\toprule
 & \multicolumn{4}{c}{MorphAlign Score} & \multicolumn{2}{c}{Overlap vs log-prob} \\
\cmidrule(lr){2-5}\cmidrule(lr){6-7}
Language & log-prob & random & reverse & PathPiece & random & reverse \\
\midrule
English & \textbf{0.57} & 0.34 & 0.20 & 0.34 & 70\% & 50\% \\
German & \textbf{0.87} & 0.56 & 0.38 & 0.58 & 74\% & 59\% \\
Finnish & \textbf{1.23} & 0.77 & 0.64 & 0.78 & 58\% & 40\% \\
\bottomrule
\end{tabular}
\caption{MorphAlign Score under three MinGram secondary-score rules, with PathPiece-BPE as a reference column. Scores are IBM1 alignment at threshold $0.01$ multiplied by $100$.``Random'' uses a fixed random per-token secondary score. ``Reverse'' uses the lowest-probability segmentation. The final two columns report the percentage of words whose tokenization is identical to log-prob for the MinGram controls. Bold indicates the highest MorphAlign Score per row.}
\label{tab:tiebreak}
\end{table}

\clearpage

\section{R\'enyi Entropy Diagnostic}
\label{app:renyi-entropy}

\Cref{tab:renyi-entropy} reports a R\'enyi-efficiency diagnostic using the definition of \citet{zouhar-etal-2023-tokenization} and the $\alpha = 3$ setting used by \citet{cognetta-etal-2024-two}: $H_3 / \log_2 |V_{\mathrm{used}}|$, where $V_{\mathrm{used}}$ is the nonzero vocabulary on the held-out evaluation corpus. This diagnostic measures how uniformly a tokenizer uses its active vocabulary, rather than how many tokens it emits. We use it only as an intrinsic probe; \citet{cognetta-etal-2024-two} show that R\'enyi efficiency is not a reliable standalone proxy for downstream quality.
BPE has the highest held-out R\'enyi-3 efficiency in every language, with ConvexTok consistently second. \mingrammi{}, MinGram and PathPiece-BPE lie in a narrow band just below these.

\begin{table}[!ht]
\centering
\small
\setlength{\tabcolsep}{3pt}
\begin{tabular}{@{}lrrrrrrr@{}}
\toprule
\multicolumn{8}{@{}l}{\textit{Held-out Goldfish Data: R\'enyi-3 efficiency $H_3 / \log_2 |V_{\mathrm{used}}|$}} \\
Method & English & German & Finnish & Russian & Arabic & Korean & Mean \\
\midrule
BPE & \textbf{0.416} & \textbf{0.430} & \textbf{0.445} & \textbf{0.406} & \textbf{0.488} & \textbf{0.468} & \textbf{0.442} \\
ConvexTok & 0.414 & \underline{0.426} & \underline{0.441} & \underline{0.404} & \underline{0.485} & \underline{0.467} & \underline{0.440} \\
MinGram & \underline{0.414} & 0.426 & 0.441 & 0.404 & 0.484 & 0.462 & 0.438 \\
PathPiece\hspace{0pt}-BPE & 0.414 & 0.425 & 0.439 & 0.403 & 0.484 & 0.462 & 0.438 \\
\mingrammi{} & 0.414 & 0.425 & 0.439 & 0.403 & 0.483 & 0.462 & 0.438 \\
FSP\hspace{0pt}-BPE\hspace{0pt}-Init & 0.412 & 0.423 & 0.436 & 0.401 & 0.480 & 0.454 & 0.434 \\
Unigram\hspace{0pt}-BPE\hspace{0pt}-Init & 0.412 & 0.423 & 0.436 & 0.401 & 0.480 & 0.453 & 0.434 \\
FSP & 0.412 & 0.422 & 0.435 & 0.402 & 0.479 & 0.451 & 0.433 \\
Unigram & 0.411 & 0.419 & 0.428 & 0.399 & 0.470 & 0.436 & 0.427 \\
\bottomrule
\end{tabular}
\caption{R\'enyi-3 efficiency for FineWeb-trained tokenizers, measured on held-out Goldfish evaluation corpora. Higher values indicate more uniform use of the nonzero vocabulary. Bold marks best-in-column; underline marks second-best.}
\label{tab:renyi-entropy}
\end{table}

\clearpage

\section{Examples of rare tokens}
\label{app:raretokens}

\Cref{tab:undertrained-token-examples} shows examples of the tokens seen
at most 718 times in the full ClimbMix training corpus, based on a frequency threshold of $10^{-7}$.
This corresponds to those shown in the rare token count column in \Cref{tab:downstream}.
Results show interesting similarities between ConvexTok, PathPiece, and \mingrammi{}, and show
that the default MinGram tokenizer resulted in \emph{zero} never-seen tokens in this cross-corpus setting,
and its 10th example falls above the threshold.
We also note that a few tokens like \tokens{\tsp{}sees} appear in the default Unigram tokenizer vocabulary,
caused by their last-minute resegmentation into \tokens{\tsp{}see,s}.

\begin{table}[h]
\centering
\small
\setlength{\tabcolsep}{3pt}
\begin{tabular}{rlllll}
\toprule
rank & BPE & FSP & FSP\hspace{0pt}-BPE\hspace{0pt}-Init & Unigram\hspace{0pt}-BPE\hspace{0pt}-Init & Unigram \\
\midrule
1 & \texttt{impse}~(6) & \texttt{\textvisiblespace{}t}~(0) & \texttt{\textvisiblespace{}s}~(0) & \texttt{\textvisiblespace{}ag}~(0) & \texttt{\textvisiblespace{}ad}~(0) \\
2 & \texttt{\textvisiblespace{}Entreprene}~(16) & \texttt{\textvisiblespace{}th}~(0) & \texttt{ings}~(0) & \texttt{ms}~(0) & \texttt{Is}~(0) \\
3 & \texttt{\textvisiblespace{}lawsu}~(17) & \texttt{res}~(0) & \texttt{\textvisiblespace{}ag}~(0) & \texttt{\textvisiblespace{}:}~(0) & \texttt{\textvisiblespace{}Its}~(0) \\
4 & \texttt{\textvisiblespace{}Palestin}~(18) & \texttt{sing}~(0) & \texttt{\textvisiblespace{}ar}~(0) & \texttt{\textvisiblespace{}af}~(0) & \texttt{\textvisiblespace{}sees}~(0) \\
5 & \texttt{senal}~(22) & \texttt{ser}~(0) & \texttt{\textvisiblespace{}ind}~(0) & \texttt{".}~(0) & \texttt{\textvisiblespace{}Complainant}~(99) \\
6 & \texttt{\textvisiblespace{}practition}~(24) & \texttt{ys}~(0) & \texttt{ts}~(0) & \texttt{”,}~(0) & \texttt{ClickFunnels}~(113) \\
7 & \texttt{\textvisiblespace{}careg}~(31) & \texttt{ses}~(0) & \texttt{\textvisiblespace{}ac}~(0) & \texttt{\textvisiblespace{}ton}~(0) & \texttt{Appellant}~(170) \\
8 & \texttt{\textvisiblespace{}lefto}~(33) & \texttt{ings}~(0) & \texttt{\textvisiblespace{}ass}~(0) & \texttt{\textvisiblespace{}ap}~(0) & \texttt{\textvisiblespace{}propecia}~(189) \\
9 & \texttt{\textvisiblespace{}inquir}~(42) & \texttt{\textvisiblespace{}au}~(0) & \texttt{\textvisiblespace{}av}~(0) & \texttt{\textvisiblespace{}Id}~(0) & \texttt{||£}~(280) \\
10 & \texttt{\textvisiblespace{}cryptocur}~(44) & \texttt{\textvisiblespace{}ed}~(0) & \texttt{ms}~(0) & \texttt{\textvisiblespace{}likes}~(0) & \texttt{\textvisiblespace{}levitra}~(330) \\
\midrule
total rare & 172 & 49 & 98 & 26 & 24 \\
\bottomrule
\end{tabular}

\medskip

\begin{tabular}{rllll}
\toprule
rank & MinGram & \mingrammi{} & PathPiece\hspace{0pt}-BPE & ConvexTok \\
\midrule
1 & \texttt{\textvisiblespace{}propecia}~(188) & {\fontencoding{T2A}\ttfamily\selectfont \textvisiblespace{}астрономия}~(0) & {\fontencoding{T2A}\ttfamily\selectfont \textvisiblespace{}астрономия}~(0) & {\fontencoding{T2A}\ttfamily\selectfont \textvisiblespace{}астрономия}~(0) \\
2 & \texttt{||£}~(280) & {\fontencoding{T2A}\ttfamily\selectfont \textvisiblespace{}Скандинавский}~(0) & {\fontencoding{T2A}\ttfamily\selectfont \textvisiblespace{}Скандинавский}~(0) & {\fontencoding{T2A}\ttfamily\selectfont \textvisiblespace{}эйтор}~(0) \\
3 & \texttt{\textvisiblespace{}levitra}~(329) & {\fontencoding{T2A}\ttfamily\selectfont \textvisiblespace{}Современная}~(1) & {\fontencoding{T2A}\ttfamily\selectfont \textvisiblespace{}Современная}~(1) & {\fontencoding{T2A}\ttfamily\selectfont \textvisiblespace{}Скандинавский}~(0) \\
4 & \texttt{\textvisiblespace{}appellant}~(386) & {\fontencoding{T2A}\ttfamily\selectfont \textvisiblespace{}направления}~(4) & {\fontencoding{T2A}\ttfamily\selectfont \textvisiblespace{}направления}~(4) & {\fontencoding{T2A}\ttfamily\selectfont \textvisiblespace{}Композиционная}~(0) \\
5 & \texttt{\textvisiblespace{}JUDGE}~(411) & \texttt{\textvisiblespace{}Complainant}~(99) & \texttt{\textvisiblespace{}Complainant}~(99) & {\fontencoding{T2A}\ttfamily\selectfont \textvisiblespace{}Практическое}~(0) \\
6 & \texttt{Funnels}~(435) & \texttt{ClickFunnels}~(113) & \texttt{ClickFunnels}~(113) & {\fontencoding{T2A}\ttfamily\selectfont \textvisiblespace{}Современная}~(1) \\
7 & \texttt{\textvisiblespace{}personals}~(445) & \texttt{\textvisiblespace{}Appellant}~(134) & \texttt{\textvisiblespace{}Appellant}~(134) & {\fontencoding{T2A}\ttfamily\selectfont \textvisiblespace{}направления}~(4) \\
8 & \texttt{\textvisiblespace{}Cialis}~(451) & \texttt{\textvisiblespace{}bitstarz}~(137) & \texttt{\textvisiblespace{}bitstarz}~(137) & \texttt{\textvisiblespace{}Negotia}~(5) \\
9 & \texttt{\textvisiblespace{}pokies}~(686) & \texttt{\textvisiblespace{}vBulletin}~(157) & \texttt{\textvisiblespace{}vBulletin}~(157) & \texttt{\textvisiblespace{}exfolia}~(16) \\
10 & \texttt{!’}~(740) & \texttt{\textvisiblespace{}Clickfunnels}~(161) & \texttt{\textvisiblespace{}Clickfunnels}~(161) & {\fontencoding{T2A}\ttfamily\selectfont \textvisiblespace{}новые}~(22) \\
\midrule
total rare & 9 & 57 & 59 & 81 \\
\bottomrule
\end{tabular}

\caption{Least common multi-unit token examples used for the `rare tokens' column in \Cref{tab:downstream}. Each cell shows the token string and full-corpus count. `total rare' shows the count of low-frequency tokens in the ClimbMix training corpus.}
\label{tab:undertrained-token-examples}
\end{table}

\end{document}